\documentclass[conference]{ieeeconf}
\IEEEoverridecommandlockouts
\pdfoutput=1
\usepackage[backend=biber,
            hyperref=true,
            url=false,
            isbn=false,
            doi=false,
            backref=false,
            style=ieee,
            citestyle=numeric-comp,
            sorting=nyt,
            block=none]{biblatex}
        
\renewcommand{\bibfont}{\small}
\usepackage{flushend}
\usepackage{balance}
\usepackage{marginnote}
\usepackage{graphics}
\usepackage[pdftex]{graphicx}
\usepackage{wrapfig}
\DeclareGraphicsExtensions{.pdf,.png,.jpg}

\usepackage[font={small}]{caption}
\usepackage{subcaption}
\usepackage[rightcaption]{sidecap}
\usepackage{pbox}

\usepackage{mathtools}
\usepackage{amsmath, amssymb, amscd}
\usepackage{ wasysym } 
\usepackage{amsfonts}
\usepackage{mathptmx} 
\usepackage{gensymb} 
\usepackage{nicefrac}       

\DeclareMathAlphabet{\mathcal}{OMS}{lmsy}{m}{n}
\DeclareSymbolFont{largesymbols}{OMX}{cmex}{m}{n}
\usepackage{textcomp} 
\usepackage{dsfont}

\usepackage{array} 
\usepackage{tabularx}
\usepackage{multirow}
\usepackage{multicol}
\usepackage{booktabs}

\usepackage[
  separate-uncertainty = true,
  multi-part-units = repeat
]{siunitx}
\usepackage[T1]{fontenc} 
\usepackage[utf8]{inputenc}
\usepackage[english]{babel} 
\usepackage{units}
\usepackage{bm}
\usepackage{times} 
\usepackage{xspace}
\usepackage{flushend}
\usepackage{csquotes}
\usepackage{makeidx}

\let\labelindent\relax
\usepackage{enumitem}

\usepackage{soul} 
\usepackage{subfiles} 

\usepackage[protrusion=true,expansion=true]{microtype}
\usepackage[yyyymmdd]{datetime}

\date{\protect\formatdate{1}{1}{2001}}

\usepackage{url}
\makeatletter
\g@addto@macro{\UrlBreaks}{\UrlOrds}
\makeatother
\usepackage{color}
\usepackage[usenames,dvipsnames,table]{xcolor}
\usepackage[pdfborder={0 0 0.5}]{hyperref}
\hypersetup{
    colorlinks=true,
    linkcolor=black,
    citecolor=black,
    filecolor=cyan,
    urlcolor=black
}

\usepackage{soul} 

\newcommand{\tocite}[1]{%
\textcolor{red}{[cite:\ifthenelse{\equal{#1}{}}{}{#1}?]}
}

\newcommand{\ignore}[1]{}

\usepackage{algorithm}
\usepackage{algpseudocode}

\newcommand{\algoName}{IRIS\xspace}
\newcommand{\algoFull}{Implicit Reinforcement without Interaction at Scale\xspace}

\begin{document}

\title{IRIS: Implicit Reinforcement without Interaction at Scale\\ 
for Learning Control from Offline Robot Manipulation Data}

\author{%
Ajay Mandlekar$^{1,2}$
Fabio Ramos$^{2,3}$
Byron Boots$^{2,4}$
Silvio Savarese$^{1}$
Li Fei-Fei$^{1}$
Animesh Garg$^{2,5}$
Dieter Fox$^{2,4}$
\thanks{ $^{1}\,$Stanford Vision \& Learning Lab, $^{2}\,$NVIDIA, $^{3}\,$University of Sydney, $^{4}\,$University of Washington, $^{5}\,$University of Toronto.
}%
}

\maketitle

\begin{abstract}
Learning from offline task demonstrations is a problem of great interest in robotics. 
For simple short-horizon manipulation tasks with modest variation in task instances, offline learning from a small set of demonstrations can produce controllers that successfully solve the task. 
However, leveraging a fixed batch of data can be problematic for larger datasets and longer-horizon tasks with greater variations. 
The data can exhibit substantial diversity and consist of suboptimal solution approaches. 
In this paper, we propose \algoFull (\algoName), a novel framework for learning from large-scale demonstration datasets. 
\algoName factorizes the control problem into a goal-conditioned low-level controller that imitates short demonstration sequences and a high-level goal selection mechanism that sets goals for the low-level and selectively combines parts of suboptimal solutions leading to more successful task completions.
We evaluate \algoName across three datasets, including the RoboTurk Cans dataset collected by humans via crowdsourcing, and show that performant policies can be learned from purely offline learning.
Additional results at \url{https://sites.google.com/stanford.edu/iris/}.
\end{abstract}


\IEEEpeerreviewmaketitle
\section{Introduction}
\label{sec:intro}





Recent research has successfully leveraged Reinforcement Learning (RL) for short-horizon robotic manipulation tasks, such as pushing and grasping objects~\cite{yu2016more,levine2016learning,fang2018learning}. However, RL algorithms face the burden of efficient exploration in large state and action spaces, and consequently need large amounts of environment interaction to successfully learn policies. Furthermore, leveraging RL for policy learning requires specifying a task-specific reward function that is often carefully shaped and crafted to assist in exploration. 

An appealing alternative to learning policies from scratch is to bring policy learning closer to the setting of supervised learning by leveraging prior experience. In Imitation Learning (IL), expert demonstrations are used to guide policy learning. The demonstrated data can be used in lieu of a reward function and also lessen the burden of exploration for the agent, ameliorating some of the aforementioned issues. However, IL has primarily been applied to small scale datasets collected by one decision maker. In order to truly reap the benefits of supervised learning, it is useful to consider how large-scale, diverse supervision can be used for task learning.


Large-scale human supervision has accelerated progress in computer vision and natural language processing~\cite{deng2009imagenet, rajpurkar2018squad2}, but policy learning has witnessed no such success. 
The advent of supervision mechanisms that allow for the collection of thousands of task demonstrations in a matter of days~\cite{mandlekar2018roboturk} motivates the following question: does a policy learning algorithm necessarily need to \textit{interact} with the system to learn a policy, or can a robust and performant policy be learned purely from external experiences provided in the form of a dataset?
For example, consider a pick-and-place task where a robot has to pick up a soda can and place it on a shelf. 
We have access to a large set of task demonstrations collected via human supervision where the soda can was placed in several initial poses and people controlled the arm to demonstrate many different approaches for grasping the can and placing it on the shelf. 
We would like to use this dataset to train a policy that can successfully solve the task.  



In order to leverage large datasets for policy learning, we argue that it is important to develop methods that are tolerant to datasets that are both suboptimal and diverse, since large-scale human supervision is likely to produce data that is highly varied in terms of both quality and task solution approaches. For example, some approaches for moving to the can and grasping it can be more efficient than others, and there are many valid ways to pick the can up. By contrast, conventional imitation learning methods assume that demonstration data is near-optimal and unimodal, and most methods start to deteriorate significantly when expert demonstrations are of lower quality, or when multiple solutions are demonstrated.

\begin{figure*}[!t]
    \centering
    \vspace{5pt}
    \includegraphics[width=0.9\linewidth]{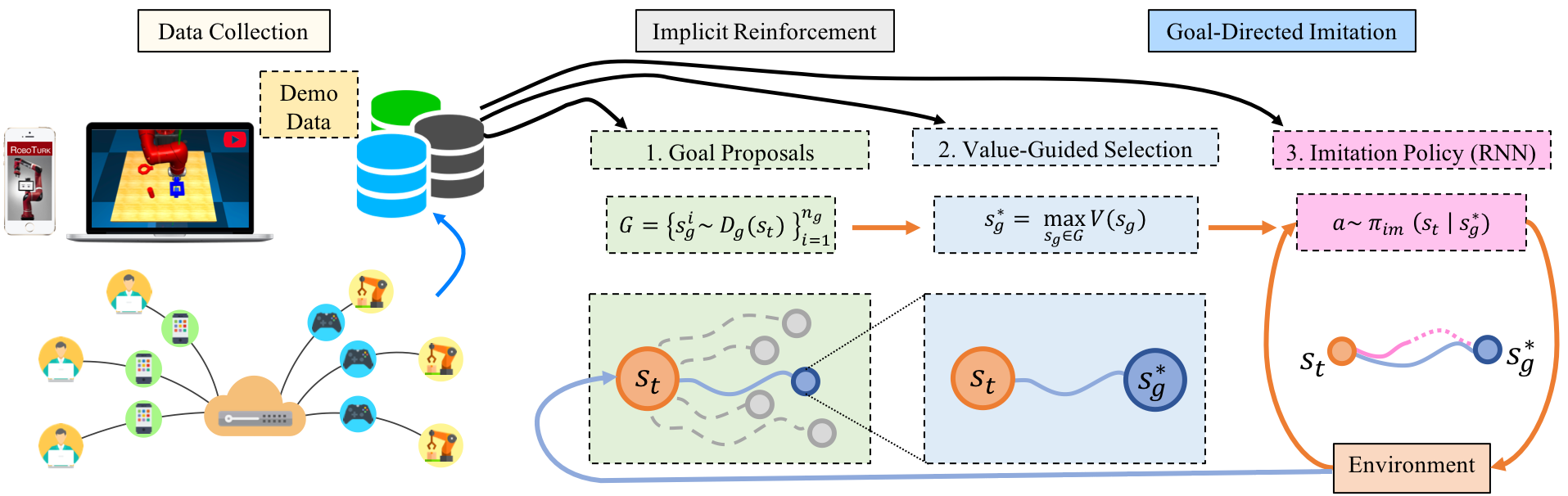}
    \caption{\textbf{Overview} \algoName learns policies from large quantities of demonstration data without environment interaction during learning. It trains a goal-conditioned low-level controller to reproduce short demonstration sequences and a high-level goal selection mechanism consisting of a goal proposal network and a value network. At test-time, a set of goals is proposed by a generative model and selected by the value function, and this is set as the target for low-level imitation. Both high and low levels are run in closed-loop with appropriate rates.}
    \label{fig:intro}
    \vspace{-20pt}
\end{figure*}

To tackle this challenge, we present \algoFull (\algoName), a novel framework that addresses the problem of offline policy learning from a large set of diverse and suboptimal demonstrations. 


\noindent \textbf{Summary of Contributions:}
\begin{enumerate}[
    topsep=0pt,
    noitemsep,
    leftmargin=*,
    itemindent=12pt]
\item We present \algoFull (\algoName), a framework that enables offline learning from a large set of diverse and suboptimal demonstrations by selectively imitating local sequences from the dataset. 
\item We evaluate \algoName across a pedagogical dataset, a highly suboptimal dataset, and a crowdsourced dataset, and only assume access to sparse task completion rewards that occur at the end of each demonstration.
\item Empirically, our experiments demonstrate that \algoName is able to leverage large-scale task demonstrations that exhibit suboptimality and diversity, and significantly outperforms other imitation learning and batch reinforcement learning baselines. 
\end{enumerate}

\section{Related work}
\label{sec:related}

\noindent \textbf{Imitation Learning and Learning from Demonstration}: Imitation learning guides policy learning by leveraging a reference set of expert demonstrations. Imitation learning methods are either offline, such as Behavioral Cloning~\cite{pomerleau1989alvinn,ross2013learning, schulman2016learning}, or online, such as Inverse Reinforcement Learning (IRL)~\cite{abbeel2011inverse,krishnan2019swirl}. Offline methods are sensitive to the quantity of demonstration data and can suffer from covariate shift, since no additional data is collected by the agent, while online methods require additional interaction for policy learning. Furthermore, most imitation learning approaches are sensitive to the quality of expert demonstrations since they assume that the data is near-optimal. \\
\noindent \textbf{Imitation and Reinforcement Learning from Suboptimal Demonstrations}: Recent work has tried to leverage off-policy deep reinforcement learning in conjunction with a set of demonstrations to account for suboptimal data and learn policies that outperform the demonstrations~\cite{zhu2018reinforcement, nac, ddpgfd, nair2018overcoming, dqfd}. However, such approaches still require significant interaction to learn policies. Furthermore, off-policy deep RL can be unstable due to the compounding effects of bootstrapping value learning and function approximation~\cite{ross2014reinforcement,fujimoto2018off, bhatt2019crossnorm, achiam2019towards}. Other methods use Batch RL to try and leverage arbitrary off-policy data for policy learning without collecting additional experience~\cite{fujimoto2018off, kumar2019stabilizing, agarwal2019striving, liu2019off, jaques2019way}. While recent efforts have produced successful continuous control policies for locomotion domains~\cite{fujimoto2018off, kumar2019stabilizing}, neither robot manipulation nor diverse demonstration data have been considered. \\
\noindent \textbf{Goal-directed Reinforcement and Imitation Learning}: Recent work has extended reinforcement learning~\cite{andrychowicz2017hindsight, pong2018temporal, nachum2018data} and imitation learning~\cite{ding2019goal, lynch2019learning} to condition on goal observations, enabling improved sample efficiency. HIRO~\cite{nachum2018data} decomposes policy learning into a high-level policy that outputs goal observations and a low-level policy that conditions on goals and tries to achieve them. While this is similar to the architecture of \algoName, our focus is on offline learning from fixed data, and our low-level policy is trained with a supervised loss similar to~\cite{lynch2019learning} instead of using an off-policy RL update, which can be unsuitable for offline learning~\cite{fujimoto2018off}. \\
\noindent \textbf{Large-Scale Data Collection in Robotics}: Self-Supervised Learning has been employed to collect and learn from large amounts of data for tasks such as grasping in both simulated~\cite{mahler2017dex,kasper2012kit,goldfeder2009columbia} and physical settings~\cite{levine2016learning,pinto2016supersizing,kalashnikov2018qt}. These methods collected hundreds of hours of robot interaction, although most of the interactions were not successful. By contrast, RoboTurk~\cite{mandlekar2018roboturk} is a platform that has been leveraged to collect large-scale datasets in simulation via crowdsourced human supervision, resulting in datasets with several successful demonstrations. We show in our experiments that \algoName can leverage such sources of demonstrations for successful policy learning without collecting additional samples of experience. 

\begin{figure*}[!t]
   \centering
   \vspace{5pt}
    \includegraphics[width =0.93\linewidth]{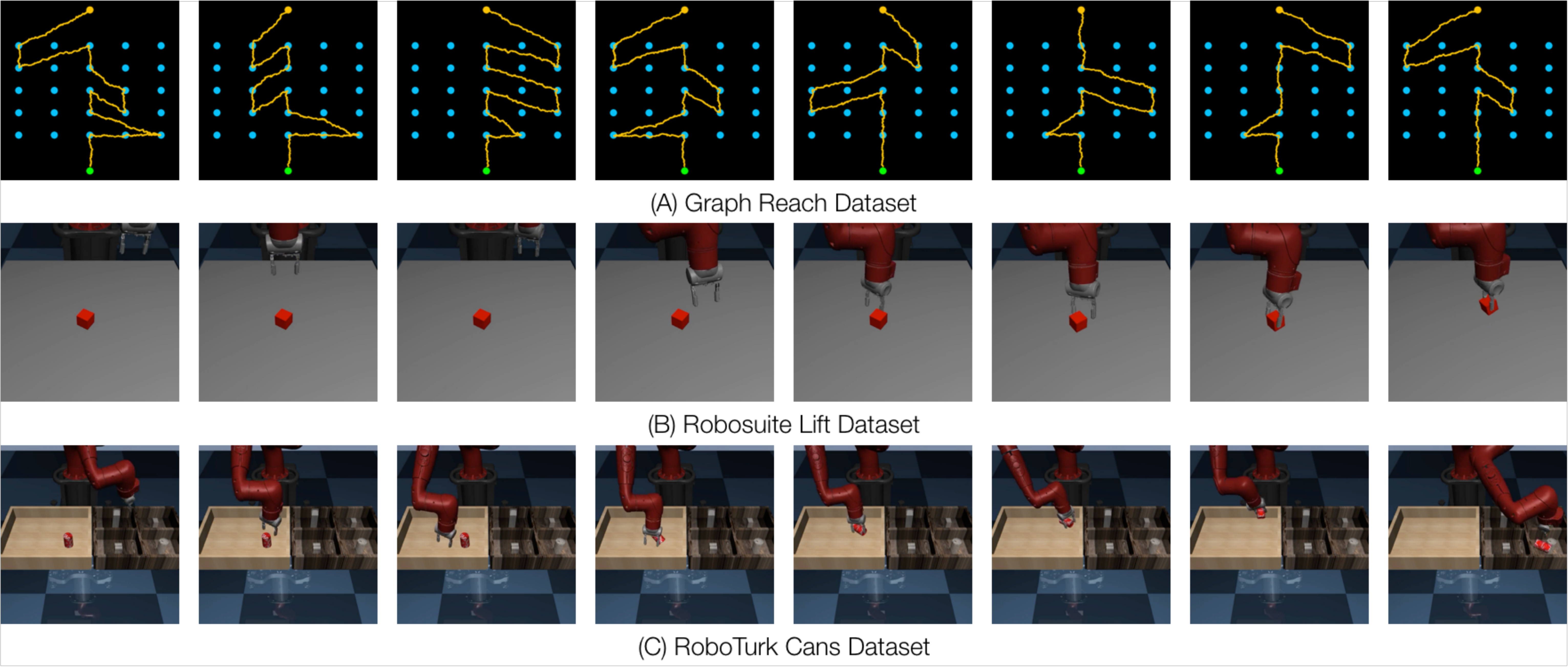}
    \caption{\textbf{Tasks and Datasets:} The Graph Reach dataset (top) consists of demonstrated paths from the start location at the top to the goal location at the bottom. Graph nodes are sampled during each demonstration to form a connected path. The Robosuite Lift dataset (middle) was collected by one human teleoperating the robot. The human intentionally took suboptimal approaches - for example, in the demonstration shown above, the robot moves close to the cube, then far away, and then fumbles with the cube before successfully lifting it. The RoboTurk Cans dataset~\cite{mandlekar2018roboturk} (bottom) was collected by several humans through crowdsourcing, leading to diverse demonstrated solutions. In the above example, the person chose to knock the can over in order to pick it up.}
   \label{fig:tasks}
   \vspace{-20pt}
\end{figure*}

\section{Preliminaries}
\label{sec:problem}
Every robot manipulation task can be formulated as a sequential decision making problem. 
Consider an infinite-horizon discrete-time Markov Decision Process (MDP) $\mathcal{M} = (\mathcal{S}, \mathcal{A}, \mathcal{T}, R, \gamma, \rho_0)$, where $\mathcal{S}$ is the state space, $\mathcal{A}$ is the action space, $\mathcal{T}(\cdot | s, a)$, is the state transition distribution, $R(s, a, s')$ is the reward function, $\gamma \in [0, 1)$ is the discount factor, and $\rho_0(\cdot)$ is the initial state distribution. At every step, an agent observes $s_t$, uses a policy $\pi$ to choose an action $a_t = \pi(s_t)$, and observes the next state $s_{t+1} \sim \mathcal{T}(\cdot | s_t, a_t)$ and reward $r_t = R(s_t, a_t, s_{t+1})$. The goal in reinforcement learning is to learn an policy $\pi$ that maximizes the expected return $\mathbb{E}[\sum_{t=0}^{\infty} \gamma^t R(s_t, a_t, s_{t+1})]$.

To use this formulation for robotic task learning, we augment this MDP with a set of absorbing goal states $\mathcal{G} \subset \mathcal{S}$, where each goal state $s_g \in \mathcal{G}$ corresponds to a state of the world in which the task is considered to be solved. Similarly, every state $s_0 \sim \rho_0(\cdot)$ corresponds to a new task instance. To measure task success, we define a sparse reward function $R(s, a, s') = \mathds{1}[s' \in \mathcal{G}]$. Consequently, maximizing expected returns corresponds to solving a task quickly and consistently. Next, we formalize the structure of the datasets we aim to leverage for task learning.

\textbf{Definition 3.1} (Goal-Reaching Trajectories) Let $\tau = (s_0, a_0, r_0, s_1, ..., s_T)$ be a $T$-length trajectory in the MDP, where $s_0 \sim \rho_0(\cdot)$ is an initial state from the MDP with rewards $r_t = R(s_t, a_t, s_{t+1})$, and states $s_{t+1} \sim \mathcal{T}(\cdot | s_t, a_t)$ produced by the MDP given the actions $a_0, a_1, ..., a_{T-1}$. This trajectory is \textit{goal-reaching} if the last state is a goal state, $s_T \in \mathcal{G}$.



In our setting, we assume access to a dataset $\mathcal{D}$ of $N$ goal-reaching trajectories that has been collected by a set of policies. Our goal is to leverage this large batch of goal-reaching trajectories to learn a policy that maximizes task returns. Importantly, the method cannot collect additional samples of experience in the MDP.
Next, we outline some dataset properties that makes learning in this setting challenging. 

\textbf{Suboptimal Data:} There are no guarantees placed on the quality of data-generating policies - each trajectory may take longer than necessary to solve the task. Equivalently, for a given trajectory in the dataset $\tau = (s_0, a_0, r_0, ..., s_T) \in \mathcal{D}$ it is possible that $\sum_{t=0}^{T-1} \gamma^t r_t + \frac {\gamma^T} {1 - \gamma} < V^*(s_0)$, so the task return of the demonstrated trajectory is worse than that of the optimal policy. 
Thus, this is different from the standard setting of imitation learning - the learned policy should not seek to imitate all demonstrated data due to variations in data quality.

\textbf{Multimodal Data:} Since many trajectories are in the dataset and multiple policies were used for generation, data can exhibit multimodality in how task instances are solved. For example, a soda can can be grasped from the top, or knocked down and then picked up on its side.


\begin{algorithm*}[!t]
\caption{\algoName: Train Loop}
\label{alg:train}
\begin{algorithmic}[1]

\Require 
\Statex $\pi_{\theta}(s \,|\, s_g)$, $\{E_{\phi}(s_g, s), D_{\phi}(z, s)\}$, $Q_{\psi}(s, a)$, $\{E_{\omega}(a, s), D_{\omega}(z, s)\}$ \Comment{Policy, Goal cVAE, Value Network, Action cVAE}
\For{$i = 1, 2, ..., n_{\text{iter}}$} 

\State $(s_t, a_t, r_t, s_{t+1}, ..., s_{t + T - 1}, a_{t + T - 1}, r_{t + T - 1}, s_{t + T}) \sim \mathcal{D}$ \Comment{Sample $T$-length sequence from the dataset}
\State $s_g \leftarrow s_{t+T}$ \Comment{Treat last observation as goal}
\State $\hat{a}_t, \hat{a}_{t+1}, ..., \hat{a}_{t + T - 1} \leftarrow \pi_{\theta}(s_{t:t+T-1} \,|\, s_g)$ \Comment{Goal-conditioned action sequence prediction from RNN Policy}
\State $\theta \leftarrow \arg\min_{\theta} \sum_{t' = t}^{t + T - 1} ||a_{t'} - \hat{a}_{t'}||_2^2$ \Comment{Update policy with imitation loss}
\State $\mu_g, \sigma_g = E_{\phi}(s_g, s_t)$, $z \sim \mathcal{N}(\mu_g, \sigma_g)$
\State $\phi \leftarrow \arg\min_{\phi} ||s_g - D_{\phi}(z, s_t) ||_2^2 + \beta_{g} KL(\mathcal{N}(\mu_g, \sigma_g) || \mathcal{N}(0, 1))$ \Comment{Train Goal cVAE to predict goals}
\State $\mu_a, \sigma_a = E_{\omega}(a_{t+T-1}, s_{t+T-1})$, $z \sim \mathcal{N}(\mu_a, \sigma_a)$
\State $\omega \leftarrow \arg\min_{\omega} ||a_{t+T-1} - D_{\omega}(z, s_{t+T-1}) ||_2^2 + \beta_{a} KL(\mathcal{N}(\mu_a, \sigma_a) || \mathcal{N}(0, 1))$ \Comment{Train Action cVAE on last action}
\State $A \leftarrow \{a_i \sim D_{\omega}(s_g) \}_{i=1}^M$ 
\State $\bar{V} = r_{t+T-1} + \gamma \max_{a_i \in A} Q_{\psi}^{'}(s_g, a_i)$ \Comment{Set target value for value update}
\State $\psi \leftarrow \arg\min_{\psi} (\bar{V} - Q_{\psi}(s_{t+T-1}, a_{t+T-1}))^2 $ \Comment{Update value network}

\EndFor
\end{algorithmic}
\end{algorithm*}





\section{\algoName: \algoFull}
\label{sec:method}

\subsection{Overview}
We first provide an overview of \algoName and motivate each component by how it is used at test-time. We split the decision making process into a high-level mechanism that sets goal states for a low-level controller to try and reach. At a state $s_t$, the high-level mechanism selects a new goal state $s_g$ that is held constant for the next $T$ timesteps. Then, the low-level controller is conditioned on $s_g$, and is given $T$ timesteps to try and reach that state in a closed-loop fashion. Then, control is returned to the high-level and the process repeats.

We further break the high-level mechanism into 2 parts. The first part is a conditional Variational Autoencoder (cVAE)~\cite{kingma2013auto} that tries to model the full distribution of states $p(s_{t+T} | s_t)$ that are $T$ timesteps away from a given state $s_t$. It is used to sample a set of goal proposals. The second part is a value function $V(s)$ that is used to select the most promising goal proposal. The low-level controller is a Recurrent Neural Network (RNN) that outputs an action $a_t$ at each timestep, given a current observation $s_t$ and goal $s_g$. 

Together, these components allow for \textit{selective imitation} of local sequences in the dataset. The complete training loop is provided in Algorithm~\ref{alg:train}. We next describe how each component is supervised.


\subsection{Low-Level Goal-Conditioned Imitation Controller}

The low-level goal-conditioned controller is a goal-conditioned RNN $\pi_{\theta}(s \,|\, g)$ (similar to~\cite{lynch2019learning}) trained on trajectory sequences of length $T$. Consecutive state-action sequences $(s_t, a_t, ..., s_{t + T - 1}, a_{t + T - 1}, s_{t + T})$ are sampled from trajectories in the dataset. The last observation in each sequence, $s_{t+T}$ is treated as a goal that the RNN should try to reach, and the RNN is trained to output the action sequence $a_t, a_{t+1}, ..., a_{t + T - 1}$ from the state sequence $s_t, s_{t+1}, ..., s_{t + T - 1}$ and the goal $s_g = s_{t+T}$ (lines 4-5 in Algorithm~\ref{alg:train}). The loss function for the RNN is a simple Behavioral Cloning loss $\mathcal{L}_{\theta}(a_{t:t+T}, s_{t:t+T}) = \sum_{k=t}^{t + T - 1}||a_k - \pi_{\theta}(s_k | s_g) ||_2^2$. By learning to copy the action sequence that resulted in a particular observation, the RNN performs unimodal imitation over short demonstration sequences to reach different goals. 

\subsection{High-Level Goal Selection Mechanism}

The high-level goal selection mechanism chooses goal states for the low-level to try and reach (similar to~\cite{nachum2018data}). The goal selection mechanism has two components: (1) a cVAE $(E_{\phi}(s_g, s), D_{\phi}(z, s))$ to propose goal states at a particular state and (2) a value function $V(s_g)$ that models the expected return of goal states. 

The cVAE is a conditional generative model that is trained on pairs of current and future observations $(s_t, s_{t + T})$ sampled from trajectories in the dataset (lines 5-7 in Algorithm~\ref{alg:train}). An encoder maps a current and future observation to the parameters of a latent Gaussian distribution $\mu_g, \sigma_g = E_{\phi}(s_{t+T}, s_t)$ and the decoder is trained to reconstruct the future observation from the current observation and a latent sampled from the encoder distribution $\hat{s}_{t+T} = D_{\phi}(z, s_t)$,  $z \sim \mathcal{N}(\mu_g, \sigma_G)$. The encoder distribution is regularized with a KL-loss $KL(\mathcal{N}(\mu_g, \sigma_g) || \mathcal{N}(0, 1))$ with weight $\beta_g$~\cite{higgins2017beta} to encourage the encoder distribution to match a prior latent distribution $p(z) = \mathcal{N}(0, 1)$ so that at test-time, the decoder can be used as a conditional generative model by sampling latents $z \sim \mathcal{N}(0, 1)$ and passing them through the decoder.

The value function consists of a state-action value function $Q_{\psi}(s, a)$ trained using a simple variant of Batch Constrained Q-Learning (BCQ)~\cite{fujimoto2018off} (lines 8-12 in Algorithm~\ref{alg:train}). 
The loss function for the value function is a modified version of the BCQ update, which maintains a cVAE $(E_{\omega}(a, s), D_{\omega}(z, s))$ to model a state-conditional action distribution $p(a | s)$ over the dataset, and a Q-network $Q_{\psi}(s,a)$ trained with a temporal difference loss, $\mathcal{L}_{\psi}(s, a, r, s') = (Q_{\psi}(s, a) - Q_{\text{target}})^2$. The target value is computed by considering a set of action proposals from the cVAE $A = \{D_{\omega}(z, s) \,|\, z \sim \mathcal{N}(0, 1) \}_{i=1}^M$ and maximizing the Q-network over the set of actions, $Q_{\text{target}} = r + \gamma \max_{a_i \in A} Q_{\psi}^{'}(s', a_i)$. 



\begin{figure*}[!t]
    \centering
    \vspace{10pt}
    \includegraphics[width =0.99\linewidth]{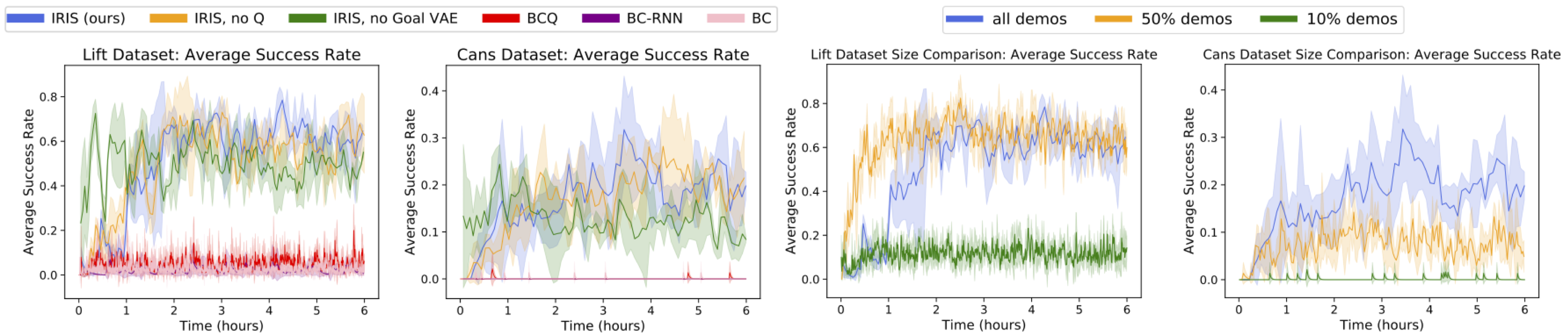}
   \caption{\textbf{Manipulation Results:} We present a comparison of \algoName against several baselines on the Robosuite Lift and RoboTurk Cans datasets (left two plots). There is a stark contrast in performance between variants of \algoName and the baseline models, which suggests that \textit{goal-conditioned imitation} is critical for good performance. We also perform a dataset size comparison (right two plots) to understand how the performance of \algoName is affected by different quantities of data.}
  \label{fig:results}
  \vspace{-15pt}
\end{figure*}

\section{\algoName: Challenges of Purely Offline Data}

In this section we elaborate on different properties of the method and how it addresses the challenges in our datasets.

\textbf{Learning from diverse solution approaches:} The goal-conditioned controller is trained to condition on future goal observations at a fine temporal resolution and produce unimodal action sequences. Consequently, it is not concerned with modeling diversity, but rather reproduces small action sequences in the dataset to move from one state to another. Meanwhile, the generative model in the goal selection mechanism proposes potential future observations that are reachable from the current observation - this explicitly models the diversity of solution approaches. In this way, \algoName decouples the problem into reproducing specific, unimodal sequences (policy learning) and modeling state trajectories that encapsulate different solution approaches (diversity), allowing for \textit{selective imitation}. 

\textbf{Learning from suboptimal data:} The low-level goal-conditioned controller operates for a small number of timesteps, so it has no need to account for suboptimal actions. This is because if the goal is to reach a state $s_2$ from $s_1$, and $T$ is sufficiently small, then a policy would only be able to improve by reaching $s_2$ in less than $T$ steps, which is a negligible improvement for small values of $T$. By contrast, the value learning component of the goal selection mechanism explicitly accounts for suboptimal solution approaches by evaluating the expected task returns of each goal and selecting the goal with the highest return. 

\textbf{Learning from off-policy datasets:} Policy learning from arbitrary off-policy data can be challenging ~\cite{fujimoto2018off, kumar2019stabilizing}. Following prior work, \algoName deals with this issue by constraining learning to occur within the distribution of training data. The goal-conditioned controller directly imitates sequences from the training data, and the generative goal model is also trained to propose goal observations from the training data. Finally, the value learning component of the goal selection mechanism mitigates extrapolation error by making sure that the Q-network is only queried on state-action pairs that lie within the training distribution~\cite{fujimoto2018off}. 


\begin{table*}[t!]
    \centering
    \vspace{4mm}
    \caption{\textbf{Performance Comparison:} We present a comparison of the best performing models for our method and baselines. Evaluations occurred on model checkpoints once per hour over 100 randomized task instances. We report the best task success rate, average rollout length (among successful rollouts), and discounted task return per training run across three random seeds. Most models are able to decrease or maintain average rollout lengths among successful rollouts compared to the original dataset of trajectories.}
    \resizebox{0.99\linewidth}{!}{%
 \begin{tabular}{| c | c c c | c c c | c c c | c c c |} 
 \hline
 \rowcolor[HTML]{CBCEFB}
  & & Graph Reach & & & Robosuite Lift & & & RoboTurk Cans & & & RoboTurk Cans Image & \\ [0.2ex] 
 \hline
  \rowcolor[HTML]{CBCEFB}
   Model & Success  & Rollout & Task &  Success & Rollout & Task & Success & Rollout & Task & Success & Rollout & Task \\ [0.2ex] 
  \rowcolor[HTML]{CBCEFB}
    & Rate (\%) & Length & Return &  Rate (\%) & Length & Return & Rate (\%) & Length & Return & Rate (\%) & Length & Return \\ [0.2ex] 
 \hline
 BC 
 & $100 \pm 0$ & $2750 \pm 346$ & $67.4 \pm 20.0$
 & $13.7 \pm 7.36$ & $404 \pm 100$ & $96.8 \pm 59.0$
 & $0.00 \pm 0.00$ & - & $0.00 \pm 0.00$
 & $13.3 \pm 4.04$ & $946 \pm 70.9$ & $55.9 \pm 17.5$
 \\
\rowcolor[HTML]{EFEFEF} 
BC-RNN 
& $100 \pm 0$ & $2918 \pm 36.1$ & $54.0 \pm 1.93$
& $16.7 \pm 10.6$ & $401 \pm 114$ & $117 \pm 75.5$
& $0.33 \pm 0.47$ & $166 \pm 235$ & $2.02 \pm 2.86$
& $28.3 \pm 1.53$ & $635 \pm 71.5$ & $157 \pm 14.8$
\\
BCQ 
& $100 \pm 0$ & $2077 \pm 162$ & $127 \pm 19.3$ 
& $18.0 \pm 13.5$ & $360 \pm 65.0$ & $132 \pm 106$ 
& $0.00 \pm 0.00$ & - & $0.00 \pm 0.00$
& $9.67 \pm 3.06$ & $706 \pm 156$ & $52.2 \pm 19.3$
\\
\rowcolor[HTML]{EFEFEF}
\algoName, no Goal VAE 
& \textbf{100 $\pm$ 0} & \textbf{1895 $\pm$ 131} & \textbf{151 $\pm$ 18.9}
& $73.0 \pm 5.35$ & $533 \pm 38.7$ & $432 \pm 47.9$
& $21.0 \pm 3.27$ & $593 \pm 15.6$ & $117 \pm 19.9$
& $38.7 \pm 6.66$ & $632 \pm 28.2$ & $213 \pm 35.1$
\\ 
\algoName, no Q 
& $100 \pm 0$ & $2285 \pm 227$ & $107 \pm 24.8$
& $74.3 \pm 14.9$ & $513 \pm 18.1$ & $447 \pm 89.4$
& \textbf{30.7 $\pm$ 3.68} & \textbf{618 $\pm$ 38.5} & \textbf{168 $\pm$ 23.8}
& \textbf{42.7 $\pm$ 5.03} & \textbf{661 $\pm$ 8.92} & \textbf{230 $\pm$ 30.2}
\\
\rowcolor[HTML]{EFEFEF} 
\algoName (Full Model) 
& $100 \pm 0$ & $2264 \pm 171$ & $106 \pm 18.4$
& \textbf{81.3 $\pm$ 6.60} & \textbf{523 $\pm$ 29.0} & \textbf{486 $\pm$ 49.7}
& \textbf{28.3 $\pm$ 0.94} & \textbf{569 $\pm$ 11.5} & \textbf{163 $\pm$ 5.68}
& \textbf{42.3 $\pm$ 1.15} & \textbf{625 $\pm$ 34.6} & \textbf{236 $\pm$ 12.3}
\\ 
\hline
Dataset (Oracle)
& $100 \pm 0$ & $3844 \pm 644$ & $27.0 \pm 22.2$
& $100 \pm 0$ & $622 \pm 192$ & $546 \pm 92.7$
& $100 \pm 0$ & $590 \pm 84.0$ & $566 \pm 48.6$
& $100 \pm 0$ & $590 \pm 84.0$ & $566 \pm 48.6$
\\
\hline
\end{tabular}
}
\label{table:performance}
\vspace{-15pt}
\end{table*}

\section{Experimental Setup}
\label{sec:exp}

\subsection{Tasks and Datasets}

\textbf{Graph Reach - A Pedagogical Example:} We constructed a simple task in a 2D navigation domain where the agent begins each episode at a start location and must navigate to a goal. The start and goal locations are fixed across all episodes. We generate a large, varied dataset by leveraging a 5x5 grid of points to sample random paths from the start location to the goal, and collecting demonstration trajectories by playing noisy, random magnitude actions to move along sampled random paths. Demonstration paths that deviate from the central path are made to take longer detours before joining the central path again (see Fig.~\ref{fig:tasks}).
Several varied demonstrated paths are available in the dataset, and only certain parts of each path should be imitated to yield optimal performance. The algorithm needs to be able to recover a policy that follows the straight line path from the start to the goal by choosing to imitate pieces of the demonstrations in the dataset (for example the first, second, and third part of the three paths respectively, in the top right 3 images of Fig.~\ref{fig:tasks}).
The dataset contains 250 demonstrations with an average completion time of 3844 timesteps. 

\textbf{Robosuite Lift - Suboptimal Demonstrations from a Human:} We collected human demonstrations from a single human using RoboTurk~\cite{mandlekar2018roboturk} on the Robosuite Lifting task~\cite{fan2018surreal}. The goal is to actuate the Sawyer robot arm to grasp and lift the cube on the table. The demonstrator lifted the cube with a consistent grasping strategy, but took their time to grasp the cube, often moving the arm to the cube and then back, or actuating the arm from side to side near the cube, as shown in Fig.~\ref{fig:tasks}. This was done intentionally to ensure that there would be several state-action pairs in the dataset with little value. Algorithms need to avoid being misled by the suboptimal paths taken by the demonstrator. The dataset contains 137 demonstrations with an average completion time of 622 timesteps.

\textbf{RoboTurk Can Pick and Place - Crowdsourced Demonstrations:} We leverage the RoboTurk pilot dataset~\cite{mandlekar2018roboturk} to train policies on the Robosuite Can Pick and Place task~\cite{fan2018surreal}. While the original dataset contained over 1100 demonstrations, we present results on a filtered version consisting of the fastest 225 trajectories. 
These demonstrations were collected across multiple humans and exhibit significant suboptimality and diversity in the solution approaches. For example, some people chose to grasp the can in an upright position by carefully positioning the gripper above the can while others chose to knock the can over before grasping the can on its side. An example of the latter is shown in Fig.~\ref{fig:tasks}. This dataset contains 225 demonstrations with an average completion time of 589 timesteps.

\textbf{RoboTurk Can Image:} This is a variant of the crowdsourced dataset that has image observations from a frontview camera instead of robot and object observations.

\subsection{Experiment Details}

We compare \algoName to a Behavioral Cloning (BC) baseline that performs simple regression over state-action pairs in the dataset, a Recurrent Neural Network (RNN) variant of Behavioral Cloning that we call BC-RNN, and a Batch-Constrained Q-Learning (BCQ) baseline, which is a state-of-the-art Batch Reinforcement Learning algorithm for continuous control~\cite{fujimoto2018off}. We also compare against two variants of \algoName to evaluate the utility of each component - a version with no Q-function at the high-level (goal selection occurs by simply sampling the Goal VAE) and a version where a deterministic goal prediction network is used in lieu of the VAE (simple regression is used to train this network). We emphasize that all training is \textit{offline} - no algorithm is allowed to collect additional samples. 



\section{Results}
\label{sec:results}

\begin{figure}[!t]
    \centering
    \vspace{5pt}
    \includegraphics[width=0.5\linewidth]{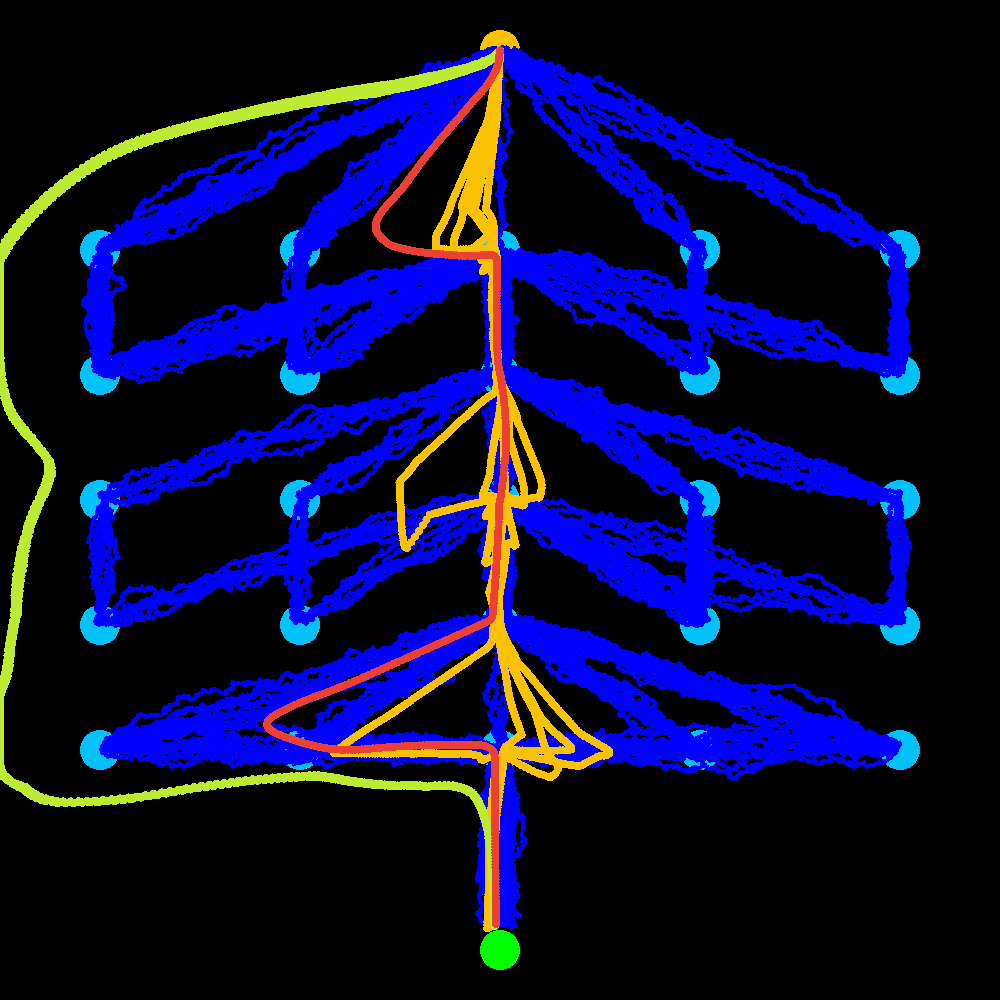}
    \caption{\textbf{Qualitative Evaluation:} We visualize 5 trajectories taken by the best performing policies for the BC (red), BCQ (green), and \algoName (orange) models on the Graph Reach environment. A set of 50 trajectories from the dataset (blue) is also shown. Our model is both able to faithfully reconstruct demonstrated trajectories and leverage them to reach the goal quickly. By contrast, BCQ extrapolates an entirely new trajectory, while BC converges to a particular mode in the dataset that is slow to reach the goal. Unlike the other models, ours also exhibits variation in policy rollouts.}
    \label{fig:qualitative}
    \vspace{-20pt}
\end{figure}


\textbf{1. Can \algoName successfully recover a performant policy by \textit{selectively imitating} pieces of a varied set of demonstrations?}
To answer this, we present quantitative results across all datasets and baselines in Table~\ref{table:performance} and also investigate qualitative model performance on the Graph Reach dataset in Fig.~\ref{fig:qualitative}. Table~\ref{table:performance} shows that while all models are able to solve the Graph Reach task consistently, variants of \algoName and BCQ are able to solve the task faster.  
To verify that \algoName is indeed imitating useful portions of demonstrated trajectories, we plot trajectories taken by the best BC (red), BCQ (green) and \algoName (orange) model in Fig.~\ref{fig:qualitative} and compare them to trajectories in the dataset (shown in blue). The plot demonstrates that our model has the capacity to imitate several different demonstrated modes from the dataset and leverage them to reach the goal quickly, while BCQ extrapolates to unseen states to reach the goal and attain similar performance. This type of extrapolation can be harmful in more complex robot manipulation tasks such as our Lift and Can tasks. This experiment demonstrates that \algoName is able to reproduce multimodal behaviors in the dataset and selectively interpolate between them to solve a task efficiently. 

\textbf{2. What benefits do our two-level decomposition provide for learning manipulation policies from diverse demonstration data?}
We consider the more challenging Lift and Cans manipulation datasets. As Table~\ref{table:performance} and the left two plots of Fig.~\ref{fig:results} show, there is a stark contrast in performance between variants of \algoName and baselines. Our models achieve success rates of 70-80\% and 20-30\% on the Lift and Cans tasks respectively while baseline models can only attain 18\% on the Lift task, and fail to solve the Cans task at all. 
The only difference between the BC-RNN model and \algoName, no Goal VAE is that \algoName conditions the RNN on goal observations and these goal observations are generated at test-time by a network that was trained to predict observations $T$ timesteps into the future. The large performance gap between these two models implies that \textit{goal-directed imitation}, which the baselines lack, is critical to deal with the multimodality in these datasets, and helps facilitate faithful imitation.

Allowing for diverse goal predictions also significantly improves performance - \algoName, no Q achieves 10\% higher success rate than \algoName, no Goal VAE on the Cans dataset by replacing a deterministic goal prediction with a VAE. 
Finally, although using the value network for goal selection did not improve performance on the Cans dataset, using value selection allowed significant improvement on the Lift task. We hypothesize that the value function helps avoid situations where the demonstrator moved away from the cube or drifted from side to side on the Lift dataset by choosing goals that lead the arm closer to the cube. 
In summary, our decomposition allows behaviors from the demonstrations to be reproduced over an extended period of time while simultaneously allowing the high-level component flexibility in dictating which behaviors should be reproduced. 


\textbf{3. How much data is necessary to train policies successfully on these tasks?}
We train \algoName on smaller subsets of the datasets - small datasets consisting of the best 10\% of the trajectories (in terms of completion time) and medium datasets consisting of the best 50\% of the trajectories. The right two plots in Fig.~\ref{fig:results} depict learning curves for \algoName on these datasets. The smaller-sized datasets lead to poor performance but the medium-sized Lift dataset has the same asymptotic performance as the full dataset. By contrast, the medium-sized Cans dataset restricts performance significantly. This shows that for tasks with greater variation in task instance, \algoName benefits from having more data in the dataset.

\textbf{4. Can \algoName train successful policies on datasets with image observations?} We train \algoName on the RoboTurk Cans Image dataset, where the observations are 128 by 128 RGB images of the robot workspace from a camera placed in front of the robot arm. We leverage the method from Dundar et al.~\cite{dundar2020unsupervised} to pre-train a landmark representation for all image observations. We use 16 landmark locations, corresponding to a 32-dimensional representation for each image. Then, we train \algoName on these representations. Surprisingly, all three IRIS variants achieve higher success rates than their counterparts on the low dimensional dataset - with the best model achieving $42.7\%$ success rate.
This result suggests that \algoName can leverage pre-trained low-dimensional representations of high-dimensional observations in order to learn performant visuomotor policies from completely offline data.

\section{Conclusion}

We introduced \algoName, a framework for offline learning from a large set of diverse and suboptimal demonstrations that operates by selectively imitating local sequences from the dataset. We demonstrated that \algoName recovers performant policies from large manipulation datasets and significantly outperforms other baselines due to our decomposition of the problem into \textit{goal-conditioned imitation} and a high-level goal selection mechanism. One limitation of the current approach is that training and testing distributions of task instances must be similar. For example, the training data must contain a sufficient variety of initial can locations to expect that the test-time policy can generalize to all can locations inside the bin. Domain adaptation for dealing with novel test-time scenarios is an exciting direction for future work.




{\footnotesize 
\section*{Acknowledgment}
\vspace{-5pt}
\renewcommand{\baselinestretch}{0.8}
Ajay Mandlekar acknowledges the support of the Department of Defense (DoD) through the NDSEG program. We thank Kevin Shih for help with training models on image observations. We thank members of the NVIDIA Seattle Robotics Lab for several helpful discussions and feedback.
}

\renewcommand*{\bibfont}{\footnotesize}
\renewcommand{\baselinestretch}{0.95} 
\begin{flushright}
\printbibliography 

@inproceedings{yu2016more,
  title={More than a million ways to be pushed. a high-fidelity experimental dataset of planar pushing},
  author={Yu, Kuan-Ting and Bauza, Maria and Fazeli, Nima and Rodriguez, Alberto},
  booktitle={Int'l Conference on Intelligent Robots and Systems},
  year={2016}
}

@inproceedings{fang2018learning,
  title={Learning Task-Oriented Grasping for Tool Manipulation from Simulated Self-Supervision},
  author={Fang, Kuan and Zhu, Yuke and Garg, Animesh and Kurenkov, Andrey and Mehta, Viraj and Fei-Fei, Li and Savarese, Silvio},
  booktitle={Robotics: Systems and Science},
  year={2018}
}

@inproceedings{deng2009imagenet,
  title={Imagenet: A large-scale hierarchical image database},
  author={Deng, Jia and Dong, Wei and Socher, Richard and Li, Li-Jia and Li, Kai and Fei-Fei, Li},
  booktitle={IEEE Conference on Computer Vision and Pattern Recognition},
%   pages={248--255},
  year={2009},
}

@article{rajpurkar2018squad2,
  title={{Know What You Don't Know: Unanswerable Questions for SQuAD}},
  author={Rajpurkar, Pranav  and Jia, Robin and Liang, Percy},
  journal={arXiv preprint arXiv:1806.03822},
  year={2018}
}

@inproceedings{fan2018surreal,
  title={SURREAL: Open-Source Reinforcement Learning Framework and Robot Manipulation Benchmark},
  author={Fan, Linxi and Zhu, Yuke and Zhu, Jiren and Liu, Zihua and Zeng, Orien and Gupta, Anchit and Creus-Costa, Joan and Savarese, Silvio and Fei-Fei, Li},
  booktitle={Conference on Robot Learning},
  year={2018}
}

@inproceedings{mandlekar2018roboturk,
  title={{RoboTurk: A Crowdsourcing Platform for Robotic Skill Learning through Imitation}},
  author={Mandlekar, Ajay and Zhu, Yuke and Garg, Animesh and Booher, Jonathan and Spero, Max and Tung, Albert and Gao, Julian and Emmons, John and Gupta, Anchit and Orbay, Emre and Savarese, Silvio and Fei-Fei, Li},
  booktitle={Conference on Robot Learning},
  year={2018}
}

@inproceedings{pomerleau1989alvinn,
  title={Alvinn: An autonomous land vehicle in a neural network},
  author={Pomerleau, Dean A},
  booktitle={Advances in neural information processing systems},
  pages={305--313},
  year={1989}
}

@inproceedings{ross2013learning,
  title={Learning monocular reactive uav control in cluttered natural environments},
  author={Ross, St{\'e}phane and Melik-Barkhudarov, Narek and Shankar, Kumar Shaurya and Wendel, Andreas and Dey, Debadeepta and Bagnell, J Andrew and Hebert, Martial},
  booktitle={Robotics and Automation (ICRA), 2013 IEEE International Conference on},
  pages={1765--1772},
  year={2013},
  organization={IEEE}
}

@incollection{schulman2016learning,
  title={Learning from demonstrations through the use of non-rigid registration},
  author={Schulman, John and Ho, Jonathan and Lee, Cameron and Abbeel, Pieter},
  booktitle={Robotics Research},
  pages={339--354},
  year={2016},
  publisher={Springer}
}

@incollection{abbeel2011inverse,
  title={Inverse reinforcement learning},
  author={Abbeel, Pieter and Ng, Andrew Y},
  booktitle={Encyclopedia of machine learning},
  pages={554--558},
  year={2011},
  publisher={Springer}
}

@article{krishnan2019swirl,
  title={SWIRL: A sequential windowed inverse reinforcement learning algorithm for robot tasks with delayed rewards},
  author={Krishnan, Sanjay and Garg, Animesh and Liaw, Richard and Thananjeyan, Brijen and Miller, Lauren and Pokorny, Florian T and Goldberg, Ken},
  journal={The International Journal of Robotics Research},
%   volume={38},
%   number={2-3},
%   pages={126--145},
  year={2019},
%   publisher={SAGE Publications Sage UK: London, England}
}

@inproceedings{zhu2018reinforcement,
  title={Reinforcement and imitation learning for diverse visuomotor skills},
  author={Zhu, Yuke and Wang, Ziyu and Merel, Josh and Rusu, Andrei and Erez, Tom and Cabi, Serkan and Tunyasuvunakool, Saran and Kram{\'a}r, J{\'a}nos and Hadsell, Raia and de Freitas, Nando and others},
  booktitle={RSS},
  year={2018}
}

@article{nac,
  title={Reinforcement learning from imperfect demonstrations},
  author={Gao, Yang and Lin, Ji and Yu, Fisher and Levine, Sergey and Darrell, Trevor and others},
  journal={arXiv preprint arXiv:1802.05313},
  year={2018}
}

@article{ddpgfd,
  title={Leveraging Demonstrations for Deep Reinforcement Learning on Robotics Problems with Sparse Rewards},
  author={Ve{\v{c}}er{\'i}k, Matej and Hester, Todd and Scholz, Jonathan and Wang, Fumin and Pietquin, Olivier and Piot, Bilal and Heess, Nicolas and Roth{\"o}rl, Thomas and Lampe, Thomas and Riedmiller, Martin},
  journal={arXiv preprint arXiv:1707.08817},
  year={2017}
}

@inproceedings{nair2018overcoming,
  title={Overcoming exploration in reinforcement learning with demonstrations},
  author={Nair, Ashvin and McGrew, Bob and Andrychowicz, Marcin and Zaremba, Wojciech and Abbeel, Pieter},
  booktitle={2018 IEEE International Conference on Robotics and Automation (ICRA)},
  pages={6292--6299},
  year={2018},
  organization={IEEE}
}

@inproceedings{dqfd,
  title={Deep q-learning from demonstrations},
  author={Hester, Todd and Vecerik, Matej and Pietquin, Olivier and Lanctot, Marc and Schaul, Tom and Piot, Bilal and Horgan, Dan and Quan, John and Sendonaris, Andrew and Osband, Ian and others},
  booktitle={Thirty-Second AAAI Conference on Artificial Intelligence},
  year={2018}
}

@article{ross2014reinforcement,
  title={Reinforcement and imitation learning via interactive no-regret learning},
  author={Ross, Stephane and Bagnell, J Andrew},
  journal={arXiv preprint arXiv:1406.5979},
  year={2014}
}

@article{fujimoto2018off,
  title={Off-Policy Deep Reinforcement Learning without Exploration},
  author={Fujimoto, Scott and Meger, David and Precup, Doina},
  journal={arXiv preprint arXiv:1812.02900},
  year={2018}
}

@article{bhatt2019crossnorm,
  title={CrossNorm: Normalization for Off-Policy TD Reinforcement Learning},
  author={Bhatt, Aditya and Argus, Max and Amiranashvili, Artemij and Brox, Thomas},
  journal={arXiv preprint arXiv:1902.05605},
  year={2019}
}

@article{achiam2019towards,
  title={Towards Characterizing Divergence in Deep Q-Learning},
  author={Achiam, Joshua and Knight, Ethan and Abbeel, Pieter},
  journal={arXiv preprint arXiv:1903.08894},
  year={2019}
}

@article{kumar2019stabilizing,
  title={Stabilizing Off-Policy Q-Learning via Bootstrapping Error Reduction},
  author={Kumar, Aviral and Fu, Justin and Tucker, George and Levine, Sergey},
  journal={arXiv preprint arXiv:1906.00949},
  year={2019}
}

@article{agarwal2019striving,
  title={Striving for Simplicity in Off-policy Deep Reinforcement Learning},
  author={Agarwal, Rishabh and Schuurmans, Dale and Norouzi, Mohammad},
  journal={arXiv preprint arXiv:1907.04543},
  year={2019}
}

@inproceedings{andrychowicz2017hindsight,
  title={Hindsight experience replay},
  author={Andrychowicz, Marcin and Wolski, Filip and Ray, Alex and Schneider, Jonas and Fong, Rachel and Welinder, Peter and McGrew, Bob and Tobin, Josh and Abbeel, OpenAI Pieter and Zaremba, Wojciech},
  booktitle={Advances in Neural Information Processing Systems},
  pages={5048--5058},
  year={2017}
}

@article{pong2018temporal,
  title={Temporal difference models: Model-free deep rl for model-based control},
  author={Pong, Vitchyr and Gu, Shixiang and Dalal, Murtaza and Levine, Sergey},
  journal={arXiv preprint arXiv:1802.09081},
  year={2018}
}

@article{ding2019goal,
  title={Goal-conditioned Imitation Learning},
  author={Ding, Yiming and Florensa, Carlos and Phielipp, Mariano and Abbeel, Pieter},
  journal={arXiv preprint arXiv:1906.05838},
  year={2019}
}

@inproceedings{nachum2018data,
  title={Data-efficient hierarchical reinforcement learning},
  author={Nachum, Ofir and Gu, Shixiang Shane and Lee, Honglak and Levine, Sergey},
  booktitle={Advances in Neural Information Processing Systems},
  pages={3303--3313},
  year={2018}
}

@article{lynch2019learning,
  title={Learning Latent Plans from Play},
  author={Lynch, Corey and Khansari, Mohi and Xiao, Ted and Kumar, Vikash and Tompson, Jonathan and Levine, Sergey and Sermanet, Pierre},
  journal={arXiv preprint arXiv:1903.01973},
  year={2019}
}

@inproceedings{goldfeder2009columbia,
  title={The columbia grasp database},
  author={Goldfeder, Corey and Ciocarlie, Matei and Dang, Hao and Allen, Peter K},
  booktitle={Robotics and Automation, 2009. ICRA'09. IEEE International Conference on},
  pages={1710--1716},
  year={2009},
  organization={IEEE}
}

@article{kasper2012kit,
  title={The KIT object models database: An object model database for object recognition, localization and manipulation in service robotics},
  author={Kasper, Alexander and Xue, Zhixing and Dillmann, R{\"u}diger},
  journal={The International Journal of Robotics Research},
  volume={31},
  number={8},
  pages={927--934},
  year={2012},
  publisher={SAGE Publications Sage UK: London, England}
}

@article{mahler2017dex,
  title={Dex-net 2.0: Deep learning to plan robust grasps with synthetic point clouds and analytic grasp metrics},
  author={Mahler, Jeffrey and Liang, Jacky and Niyaz, Sherdil and Laskey, Michael and Doan, Richard and Liu, Xinyu and Ojea, Juan Aparicio and Goldberg, Ken},
  journal={arXiv preprint arXiv:1703.09312},
  year={2017}
}

@inproceedings{levine2016learning,
  title={Learning hand-eye coordination for robotic grasping with large-scale data collection},
  author={Levine, Sergey and Pastor, Peter and Krizhevsky, Alex and Quillen, Deirdre},
  booktitle={ISER},
  pages={173--184},
  year={2016},
}

@inproceedings{pinto2016supersizing,
  title={Supersizing self-supervision: Learning to grasp from 50k tries and 700 robot hours},
  author={Pinto, Lerrel and Gupta, Abhinav},
  booktitle={Robotics and Automation (ICRA), 2016 IEEE International Conference on},
  year={2016},
  organization={IEEE}
}

@article{kalashnikov2018qt,
  title={Qt-opt: Scalable deep reinforcement learning for vision-based robotic manipulation},
  author={Kalashnikov, Dmitry and Irpan, Alex and Pastor, Peter and Ibarz, Julian and Herzog, Alexander and Jang, Eric and Quillen, Deirdre and Holly, Ethan and Kalakrishnan, Mrinal and Vanhoucke, Vincent and others},
  journal={arXiv preprint arXiv:1806.10293},
  year={2018}
}

@article{kingma2013auto,
  title={Auto-encoding variational bayes},
  author={Kingma, Diederik P and Welling, Max},
  journal={arXiv preprint arXiv:1312.6114},
  year={2013}
}

@article{higgins2017beta,
  title={beta-VAE: Learning Basic Visual Concepts with a Constrained Variational Framework.},
  author={Higgins, Irina and Matthey, Loic and Pal, Arka and Burgess, Christopher and Glorot, Xavier and Botvinick, Matthew and Mohamed, Shakir and Lerchner, Alexander},
  journal={ICLR},
  volume={2},
  number={5},
  pages={6},
  year={2017}
}

@article{liu2019off,
  title={Off-Policy Policy Gradient with State Distribution Correction},
  author={Liu, Yao and Swaminathan, Adith and Agarwal, Alekh and Brunskill, Emma},
  journal={arXiv preprint arXiv:1904.08473},
  year={2019}
}

@article{jaques2019way,
  title={Way Off-Policy Batch Deep Reinforcement Learning of Implicit Human Preferences in Dialog},
  author={Jaques, Natasha and Ghandeharioun, Asma and Shen, Judy Hanwen and Ferguson, Craig and Lapedriza, Agata and Jones, Noah and Gu, Shixiang and Picard, Rosalind},
  journal={arXiv preprint arXiv:1907.00456},
  year={2019}
}

@article{dundar2020unsupervised,
  title={Unsupervised Disentanglement of Pose, Appearance and Background from Images and Videos},
  author={Dundar, Aysegul and Shih, Kevin J and Garg, Animesh and Pottorf, Robert and Tao, Andrew and Catanzaro, Bryan},
  journal={arXiv preprint arXiv:2001.09518},
  year={2020}
}
\end{flushright}

\end{document}